\documentclass[letterpaper]{article} 
\usepackage{aaai25}  
\usepackage{times}  
\usepackage{helvet}  
\usepackage{courier}  
\usepackage[hyphens]{url}  
\usepackage{graphicx} 
\usepackage{natbib}  
\usepackage{caption} 
\usepackage{amsmath}
\usepackage{amssymb}
\usepackage{subcaption}
\usepackage{tablefootnote}

\title{Emergent AI Surveillance: Overlearned Person Re-Identification and Its Mitigation in Law Enforcement Context}
\author {
    An Thi Nguyen\textsuperscript{\rm 1},
    Radina Stoykova\textsuperscript{\rm 2},
    Eric Arazo\textsuperscript{\rm 3}
}
\affiliations {
    \textsuperscript{\rm 1}Norwegian University of Science and Technology\\
    \textsuperscript{\rm 2}University of Groningen\\
    \textsuperscript{\rm 3}CeADAR Ireland's Centre for AI\\
    an.t.nguyen@ntnu.no, r.stoykova@rug.nl, eric.arazo@ucd.ie
}

\begin{document}

\maketitle

\begin{abstract}
Generic instance search models can dramatically reduce the manual effort required to analyze vast surveillance footage during criminal investigations by retrieving specific objects of interest to law enforcement. However, our research reveals an unintended emergent capability: through overlearning, these models can single out specific individuals even when trained on datasets without human subjects. This capability raises concerns regarding identification and profiling of individuals based on their personal data, while there is currently no clear standard on how de-identification can be achieved. We evaluate two technical safeguards to curtail a model’s person re-identification capacity: index exclusion and confusion loss. Our experiments demonstrate that combining these approaches can reduce person re-identification accuracy to below 2\% while maintaining 82\% of retrieval performance for non-person objects. However, we identify critical vulnerabilities in these mitigations, including potential circumvention using partial person images. These findings highlight urgent regulatory questions at the intersection of AI governance and data protection: How should we classify and regulate systems with emergent identification capabilities? And what technical standards should be required to prevent identification capabilities from developing in seemingly benign applications?
\end{abstract}

%
\begin{links}
    \link{Code}{https://github.com/AlpakkAn/overlearned-person-reid}
\end{links}

\section{Introduction}

\label{introduction}
Computer vision is a field of computer science that explores methods to enable computers to achieve an understanding of visual data like images or videos.
These technologies offer substantial benefits to law enforcement agencies (LEAs) by dramatically reducing the manual effort required to analyze vast surveillance footage during investigations. They have a wide range of applications, from the detection of weapons and graffiti to facial recognition. Object detection is, for example, used by Europol to detect objects in the background of images containing sexually explicit material involving minors and, based on crowdsourcing, to collect information about the origin and location of these objects to potentially identify crime perpetrators and victims \cite{Europol}.

Instance search is a task that is particularly useful in criminal investigations and aims to retrieve a specific object in an image or video collection given one or multiple visual examples of the object \citep{awad2017instance}. Then, \textit{generic} instance search refers to the capability of searching for a variety of different object classes \cite{tao2015attributes}. 
While tasks such as object detection focus on localizing and classifying all objects belonging to some defined classes (e.g. detecting all cars in the videos), instance search retrieves all occurrences of a specific object (e.g. finding every appearance of a particular car).
Person re-identification (re-ID) \footnote{ Throughout this paper, ``person re-identification" refers to the computer vision task of matching images of the same individual across camera views. This differs from ``re-identification" in privacy literature, which broadly refers to any process linking anonymized data back to individuals. For brevity, we often shorten ``person re-identification" to ``re-identification" or ``re-ID".} can be viewed as a special case of instance search. The aim of person re-ID is to associate an individual's identity across a network of cameras within a collection, typically consisting of sequences of images or video.
The individual in question, or \textit{query person}, is most commonly depicted through a still image, but can in some applications also be a sequence of video frames, textual description \citep{shao2022learning, yan2023clip}, sketch \citep{lin2023beyond, chen2023towards}, or other representations of a person.

While person re-ID offers valuable capabilities for law enforcement, these methods trigger concerns regarding personal data protection of individuals. More concerning, however, is that AI systems can acquire the ability to derive personal information from video recordings and single out concrete individuals without being designed as identification systems, which can be privacy-intrusive and result in subtle forms of surveillance.

\begin{figure*}[h]
\centering
\includegraphics[width=\linewidth]{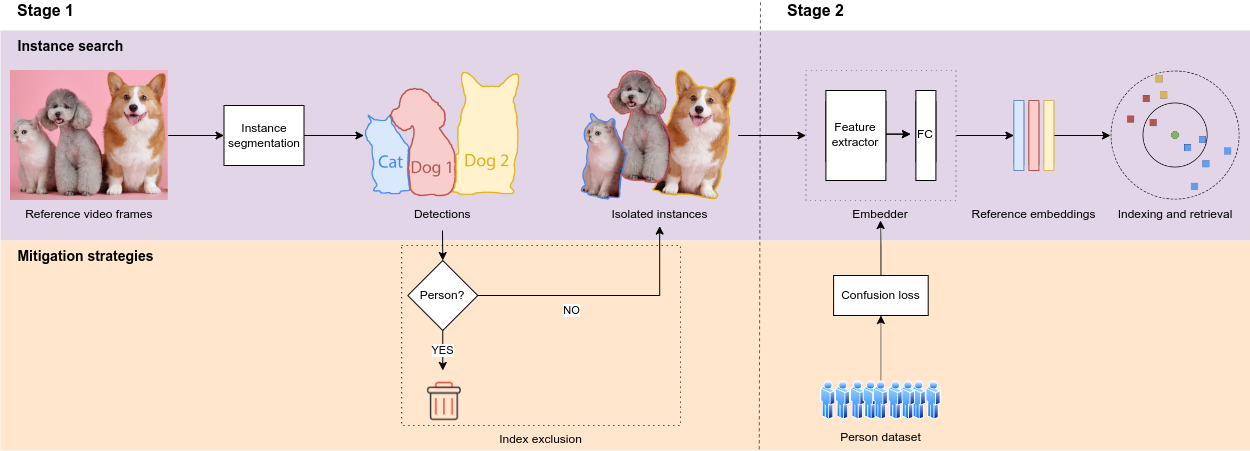}
\caption{Instance search pipeline with mitigation strategies, adapted from \citet{10.1145/3617233.3617249}.}
\label{fig:architecture}
\end{figure*}

Even before the rise of modern machine learning to derive complex patterns from data, \citet{ohm2009broken} revealed how supposedly anonymized datasets can be deanonymized by linking them to auxiliary information, posing significant privacy risks. With the advent of modern computer vision methods, \citet{Dietlmeier} showed that face anonymization provides minimal protection in person re-ID systems. Our work reveals that these challenges are only amplified as generic instance search models trained exclusively on non-human data still develop person re-ID capabilities, a phenomenon introduced by~\citet{song2020overlearning} as overlearning.
This demonstrates that privacy-invasive capabilities can develop as emergent properties of the learning process itself, regardless of data anonymization efforts.
To address these emergent privacy risks, we propose two complementary technical safeguards: index exclusion filters out persons before they are stored in the searchable database, and confusion loss intervenes during training to make representations of the same person appear dissimilar, disrupting the system's ability to consistently recognize individuals.
However, our analysis reveals that these safeguards remain vulnerable to circumvention through partial-region matching strategies and may exhibit demographic biases that unevenly protect different groups, highlighting the ongoing challenge of developing robust and equitable privacy protections in AI systems.

This paper addresses the following research questions:
\label{researchquestion}
\begin{itemize}
    \item Can instance search models develop person re-identification abilities even when trained  on datasets without human subjects?
    \item If so, what technical safeguards can effectively mitigate these overlearned capabilities?
\end{itemize}
These questions raise regulatory and technical challenges at the intersection of AI governance and data protection, which, although briefly addressed, need to be researched further. The main contributions of this work are as follows:
\begin{itemize}
    \item \textbf{Demonstration of emergent person re-ID capabilities:} We show that generic instance search models develop significant person re-ID abilities even when trained exclusively on datasets without human subjects, achieving up to 87.2\% mAP across multiple model architectures.
    \item \textbf{Technical safeguards:} We propose and evaluate two complementary approaches to curtail overlearned person re-ID capabilities: (1) index exclusion, which filters out person embeddings during indexing, and (2) confusion loss, which disperses person embeddings in the feature space during training.
    \item \textbf{Comprehensive evaluation of safeguards:} We demonstrate that combining index exclusion with confusion loss can reduce person re-ID accuracy to below 2\%, with non-person retrieval performance dropping by only 8-12 percentage points across datasets.
    \item \textbf{Vulnerability analysis:} We identify critical weaknesses in the proposed safeguards, showing that person re-ID can be partially restored using cropped person regions, clothing items, or accessories, highlighting the need for more robust defenses.
    \item \textbf{Cross-dataset validation:} We provide extensive experimental validation across multiple datasets (YouTube-VIS, OVIS, CUHK03, Market-1501) and model architectures (CLIP, ResNet-152, CAFormer, EVA-02) to demonstrate the generality of the overlearning phenomenon, and evaluate our mitigation strategies across multiple datasets using CLIP.
    \item \textbf{Legal analysis:} We briefly examine the implications of emergent identification capabilities on the European data protection regime, showing that the current regulatory framework is ambiguous about the increasing AI surveillance with re-ID systems that acquire biometric-like capabilities without explicit design.
\end{itemize}

This paper is structured as follows: First, we provide a brief overview of person re-ID and its data protection implications, establishing the regulatory context for our investigation. Next, we identify the concept of overlearning in instance search models and propose several technical mitigation strategies to curtail overlearned person re-ID capabilities. Our experimental evaluation demonstrates that person re-identification is indeed possible without human training data and is followed by a comprehensive assessment of our proposed mitigation strategies. 
Finally, we discuss the broader implications of these findings.

\section{Technical Background}
\label{sec:background}

This section describes the foundational topics underlying our investigation of emergent person re-ID capabilities in generic instance search models. We begin by describing the instance search pipeline, followed by an overview of person re-ID systems and the phenomenon of overlearning in neural networks.

\subsection{The Instance Search Pipeline}

Figure \ref{fig:architecture} illustrates the two-stage instance search pipeline we use, adapted from \citet{10.1145/3617233.3617249}. This architecture is designed to solve a practical challenge: while we need systems that can search for many different types of objects, there are not enough comprehensive datasets specifically built for this purpose. The two-stage approach leverages existing video datasets to train more capable models.

\subsubsection{Stage 1: Object Isolation}
Instance segmentation separates individual objects from backgrounds, creating focused images of each object. This allows the system to analyze object characteristics without background distractions.
\subsubsection{Stage 2: Creating Searchable Representations}
Objects are transformed into ``embeddings", which are numerical fingerprints that capture their visual characteristics. The system learns to create these embeddings such that similar objects (multiple views of the same car) have similar representations, while different objects have distinct representations.
The embedding model is trained using the multi-similarity (MS) loss \cite{wang2019multi}. For each anchor $i$, we identify challenging examples: ``hard" positives $\mathcal P_i$ (same-class samples that are difficult to recognize as similar) and ``hard" negatives $\mathcal N_i$ (different-class samples that might be mistakenly considered similar). The MS loss combines two objectives:
\begin{align}
\mathcal L_{\mathrm{MS}}
&= \frac{1}{m}\sum_{i=1}^{m}\Bigg[
\frac{1}{\alpha}\log\!\Bigl(1+\!\!\sum_{j\in\mathcal P_i} e^{-\alpha\,(s_{ij}-b)}\Bigr)  \notag \\
&\quad +\frac{1}{\beta}\log\!\Bigl(1+\!\!\sum_{j\in\mathcal N_i} e^{\beta\,(s_{ij}-b)}\Bigr)
\Bigg], \label{eq:ms}
\end{align}
where $s_{ij}=\cos(\mathbf{f}_i,\mathbf{f}_j)$ represents the cosine similarity between embeddings $\mathbf{f}_i$ and $\mathbf{f}_j$.
The first term pulls hard positives together while the second pushes hard negatives apart, with $\alpha$, $\beta$, and $b$ controlling how aggressively the model performs this separation.
Once embeddings are generated, they are stored in a searchable index --- a data structure that enables efficient retrieval of similar instances. Crucially, in most instance search systems, an object only becomes searchable once its embedding is added to this index, making the indexing step a natural checkpoint for controlling what the system can find.

\subsection{Person Re-Identification Systems}

Person re-ID represents a specialized application of instance search focused on associating individuals across multiple camera views or temporal instances. Similar to the two-stage instance search pipeline, the person re-ID pipeline begins with object isolation. First, person detection or tracking algorithms produce bounding boxes that segment individuals from their backgrounds \citep{ye2021deep}. In the second stage, consistent with the embedding creation step outlined in the instance search pipeline, person re-ID systems transform each isolated individual into discriminative embeddings or features. By ensuring that embeddings of the same individual across different images remain highly similar, the model effectively supports identity matching.

Historically, feature extraction methods in person re-ID were divided into two approaches: handcrafted features that relied on color, texture \citep{liu2012person}, or interest point matching \citep{khedher2015local}, and learned representations derived from data. Early methods often relied on hand-crafted features, but the advent of deep learning techniques has shifted the paradigm significantly towards learned representations \citep{bengio2013representation}. Representation learning techniques, such as those using MS loss or similar functions, are now predominantly utilized for their ability to discover meaningful patterns directly from data, enhancing accuracy and robustness.

Person re-ID systems also vary by application context, broadly categorized into short-term and long-term re-ID. Short-term re-ID assumes consistent appearance across multiple camera views within brief intervals, which is typical in controlled environments like airports or train stations. In contrast, long-term re-ID considers significant variability in appearance over extended periods, addressing scenarios where individuals might change clothing or accessories, making the task considerably more challenging \citep{qian2020long}.

While person re-ID systems are explicitly designed to identify and track individuals across camera networks, our investigation reveals a more concerning phenomenon: generic instance search models trained exclusively on non-human objects can spontaneously develop similar re-identification capabilities. This emergent behavior --- where models distinguish between individual persons despite never seeing human training data --- exemplifies how neural networks acquire capabilities beyond their intended objectives, a concept known as overlearning, which has significant privacy implications.

\subsection{The Overlearning Phenomenon}
The concept of \textit{overlearning} was introduced in \citet{song2020overlearning} to describe how neural networks unintentionally acquire capabilities beyond their training objectives, particularly the ability to recognize privacy-sensitive attributes not present in their learning targets. Their research demonstrated that this phenomenon appears intrinsic to deep learning systems, suggesting that certain learning objectives inherently require models to develop representations that capture generic features useful for multiple tasks, including those with privacy implications. Their work established that model representations can leak sensitive information and demonstrated that censoring techniques often fail to prevent this leakage without significantly degrading model performance \cite{song2020overlearning}. 
While their investigation focused on proving that sensitive attributes could be extracted using auxiliary models, our work extends this understanding by demonstrating that person re-identification capabilities specifically emerge in generic instance search models even without human-focused training data. Additionally, where Song and Shmatikov found limited success with censoring approaches, we evaluate multiple technical mitigation strategies, including a confusion loss technique that potentially offers a more balanced privacy-accuracy trade-off.

\subsection{Machine Unlearning}
Machine Unlearning (MU) enables removing specific data influence from models, crucial for privacy regulations like the ``right to be forgotten" \cite{wang2024machine}. Gradient-based unlearning methods modify the model's parameters using gradients that increase loss on the forget set while regularizing with retain set signals to preserve utility \cite{huang2024unified}. 
Some methods leverage knowledge distillation, where an unlearned model is trained to mimic the behavior of the original model on the retain set, while simultaneously being discouraged from matching the original model's behavior on the forget set \cite{kurmanji2023towards}.
SCAR is an approximate unlearning method that avoids using a retain set by employing a modified Mahalanobis distance to guide unlearning and a distillation-based approach with out-of-distribution images to maintain model performance. Their work also proposes a self-forget version of SCAR capable of unlearning without access to the forget set \cite{bonato2024retain}.
While MU methods predominantly target eliminating the influence of specific training examples from a model's learned parameters, they require considerable adaptation for removing overlearned capabilities.

\section{Data Protection Challenges in Instance Search Systems}
\label{sec:data_protection}
The overlearning phenomenon in instance search models creates a challenge in the European data protection regime: these systems acquire capabilities to single out individuals without explicit design intent. 
This section examines the legal status of overlearned person re-ID capabilities under the Law Enforcement Directive \cite{led_directive_2016} to assess: (1) when instance search systems trigger personal data processing obligations, (2) whether such capabilities constitute profiling under Art. 11 LED, and (3) how soft biometric features in low-resolution embeddings relate to the legal definition of biometric data.

\subsubsection{Personal Data}

Instance search models are designed to assist with the analysis of video surveillance recordings which contain visual representations of individuals. ECJ has clarified that images of persons recorded on video cameras are personal data \cite[para 22]{Rynes}  \cite[para 32]{Buivids}, if they ``can be linked to a particular person" \cite{Nowak}. An argument can be made that if the system is used for the identification of a few concrete suspects, only to them such a link can be established, while for the majority of people the video recordings are not personal data because they were never singled out or identified. This argument is rejected by the Article 29 Working Party (WP29), who argued that as “the purpose of video surveillance is to identify the persons to be seen in the video images […] the whole application as such has to be considered as processing data about identifiable persons, even if some persons recorded are not identifiable in practice” \cite{art29}.
\citet{tosoni_definitions_2022} expand on this view arguing that the mere capture of individuals' visual representation falls under the data protection regime even when such capture does not aim at identification and irrespective of whether it is performed only for a short period of time (short-term re-ID).

Law enforcement agencies seeking privacy-compliant solutions can deploy instance search models trained exclusively on non-human object datasets, avoiding person-class data entirely. As our experiments demonstrate, training instance search models on non-personal data does not prevent these algorithms from developing overlearned person re-ID capabilities. This means that such AI systems can learn to identify similar objects in a privacy-aware manner during training. However, once deployed, the algorithm exhibits high adaptability: when presented with a dataset containing people, it will detect the person as an object (although not one seen before) and generate embeddings that enable re-identification across all instances in the dataset. In this regard, the EDPB clarifies that even when ``an AI model has not been intentionally designed to produce information relating to an identifiable natural person", if personal data can be obtained via statistical or prompt interfaces, such AI model is not considered anonymous and therefore the data protection regime applies (EDPB, \citeyear{art29AI}). Applied to instance search, this means that although a model can be trained as an ``anonymous" AI system, using a deployment dataset that contains personal data will still trigger data protection obligations for the deployer.

\subsubsection{Profiling}
As re-ID techniques always process personal data when deployed to a dataset with visual representations of persons, they can be further considered under the legal regime of profiling. Art. 11 LED states that a decision based solely on automated processing, including profiling, that allows law enforcement to evaluate personal aspects of individuals and produce adverse legal effects or significantly affects them, is only permissible if: \textit{(i)} authorized by Union or Member State law; \textit{(ii)} with appropriate safeguards for individuals' rights; and \textit{(iii)} must not result in discrimination. Re-ID techniques in instance search can detect a person's visual representation across large datasets which also provides context on their locations and behaviors e.g., deriving suspects' escape path, acquaintances, cars, or accessories. Detecting objects of interest to the investigation can further point to suspects, witnesses, or victims. However, such profiling capabilities depend on how often a person appears and is detected in a video dataset. If someone is detected in 2-3 instances around a crime scene this might be sufficient to raise suspicion. The accumulation of more detections might also disclose sensitive data e.g., if a person is often detected entering religious or political buildings (EDPB, \citeyear{EDPBprofiling}). Instance search can also detect a person's visual representation based on textual description, where such description can be considered a ``profile" for the person applied to the dataset. 

However, while instance search of persons may assist in profiling, it is unlikely that it will satisfy the legal threshold for automated decision-making in
Art. 11 LED. The provision requires that individuals suffer adverse legal or significant effects based solely on algorithmic profiling. In GDPR context, it has been argued that algorithms trained to classify people in lists as suspicious or not (e.g., security screening) \cite{binns_is_2021} or creditworthy or not \cite{Schufa} produce an automated decision that impacts the individual. By contrast, instance search has no such capabilities to evaluate individuals. Even if the algorithm assists in detecting a suspect on a video, this is rather used internally by law enforcement and corroborated with other facts of the investigation, without necessarily resulting in investigative actions against an individual and therefore producing legal effects for her. Consequently, even when visual cues might be used by investigators to create a profile of a suspect, this should be considered a preparatory profiling activity and therefore out of Art.11 LED scope.  

\subsubsection{Biometric Data}

The legal regime of biometrics processing for law enforcement purposes does not clearly address instance search re-ID capabilities either. According to Art. 3 (13) LED, biometric data is a special category of personal data (1) resulting from specific technical processing (2) relating to the physical [e.g., face, voice, fingerprint], physiological [e.g., DNA] or behavioral [e.g. gait] characteristics of a natural person, (3) which allow or confirm the unique identification of that natural person. Biometrics are special categories of personal data as they are interpreted as ``stable at least for considerable periods of a human’s life, [...] distinctive if not unique for each individual [...and] allowing a more accurate way of distinguishing one person from another" \cite{kuner_article_2020}.

``Specific technical processing" is interpreted as creating measures of biometric characteristics and developing a biometric template in a reference database with key features from the raw form of biometric data (e.g. facial measurements from an image) (EDPB, \citeyear{EDPBSoftbio}). If instance search models are prompted with a photo of a suspect (input), the system follows the same procedure as biometric systems, described as creating a template from a photo and comparing it with the visual representation of a person in the CCTV recordings to output all matches \cite[p.212]{kuner_article_2020}.

However, most re-ID methods are trained and evaluated on low-resolution datasets that often do not capture distinct biometric features \citep{gawande2020pedestrian, wu2019deep}. The person is matched not based on ``primary" biometrics as those described in the legal definition, but rather on so-called soft biometrics, e.g. skin tone, hair color, tattoo, scars, body shape, clothes, and accessories \cite{dantcheva_what_2016}. It is often opaque which cues the deep learning‑based instance search model actually relies on at match time. We refer to these opaque, learned features as ``latent soft biometric representations" --- continuous embeddings that may correlate with traditional soft biometric attributes but are not explicitly extracted or labeled as such by the model. Empirically, in any given match it may be unclear whether the system used soft biometrics (e.g. body shape) or soft cues (e.g. clothing/accessories), which only in combination lead to re-identification. 
Our results demonstrate that clothing and accessories alone account for substantial re-identification capability. When queries and references were restricted to upper clothing regions, our models achieved 25.6\% mAP on CUHK03 (compared to 33.8\% mAP for full-person images), while bag-based queries yielded 18.5\% mAP (Table \ref{table:robustness}). This aligns with findings from \citet{Dietlmeier}, who showed that faces are not necessary for effective person re-identification. 
However, when a re-ID system relies on clothing, performance drops sharply when outfits change \cite{li2021learning}. Yet despite relying on characteristics that are neither stable over time, nor distinctive or unique for individuals, in combination these features enable concerning levels of identification capability.

According to EDPB, soft biometrics are traits that are not \textit{per se} personal data or biometrics as each of them does not uniquely identify a person \cite[p.16]{WP29Softbio}. Further, EDPB states that classifying people by age and gender from a video, without generating biometric templates to uniquely identify anyone, would not fall under processing of special categories of data (Art. 9 GDPR), ``as long as no other types of special categories of data are being processed" (EDPB, \citeyear{EDPBSoftbio}, p.~19). This argument clearly shows that processing of only two soft biometric features doesn't trigger Art.9 GDPR. However, some soft biometrics are not specific for a person but can partly show sensitive data, e.g. skin color is related but not directly revealing ethnicity. The EDPB guidelines do not establish whether the accumulation of more soft biometrics, in the absence of concrete special categories of data, can also trigger Art. 9 GDPR. But the lack of guidance on soft biometrics by the legislator makes it difficult to consider that unique biometric identification encompasses identification based on many latent soft biometrics.  

Further, the meaning of the term ``unique identification" in the context of biometrics has been a subject of discussion in the legal literature \cite{jasserand2016legal, jasserand2022biometric, kindt2018having}. \citet{jasserand2016legal} interprets that not all images will be regarded as biometric data, but only those which are processed to ``allow the unique identification'' of a person. For the example of facial images, criteria such as lighting, exposure, and camera resolution need to be considered \cite{jasserand2016legal}. In technical terms, this is referred to as distinctive and stable over time features \cite{jasserand2022biometric}. Distinctiveness implies that biometric data derived from different individuals should have high variation \cite{iso}. 

In our view, re-ID systems are unlikely to fall under the legal regime of biometric identification. They do not fulfill the legal criteria of ``biometric characteristics" and ``unique identification", as they use latent soft biometrics that are not stable, distinctive or unique to individuals, while the models lack primary biometric recognition functionalities. If we interpret the definition of biometric identification expansively as including the accumulation of latent representations of clothes, accessories, and body shapes, this will mean that any AI-based search of a person's visual representation is biometric identification. Such an expansion will blur the line between biometric and personal data. The inconsistencies in the legal regime of soft biometrics and the functional creep of their latent representations in AI systems raise concerns about how law enforcement can comply with data protection principles like purpose limitation and data minimization when AI systems are not trained or designed for the purpose of establishing the identity of a natural person but develop unforeseen data processing capabilities post-deployment. While further legal research needs to address these questions, this paper focuses on identifying overlearned person re-ID capabilities in instance search models and evaluating technical safeguards to mitigate them, enhancing the privacy-aware use of such algorithms. 

\section{Technical Safeguards: Mitigating Overlearned Person Re-ID}
\label{sec:experiments}
In this section, we introduce two technical mitigation strategies to curtail overlearned capabilities in instance search models.
These strategies represent distinct approaches: a practical post-processing filter (index exclusion) and an embedding space manipulation technique (confusion loss).

\subsection{Index Exclusion}

Index exclusion is a straightforward strategy: during indexing, instances classified as ``person" via instance segmentation or object detection are identified, and their feature embeddings are deliberately omitted from the search index, preventing their retrieval.

However, this method's effectiveness strongly depends on the accuracy of the initial object detection or instance segmentation step. If individuals are misclassified, their embeddings \textit{are} indexed, which undermines the strategy. This is particularly concerning given known algorithmic biases in computer vision. Research documents significant performance disparities across demographic groups in relevant tasks like object and pedestrian detection \citep{li2025bias, wilson2019predictive, gustafson2023facet}, influenced by factors such as skin tone, age, and gender combined with environmental conditions \citep{li2025bias}.
If the detection model exhibits such biases, systematically failing to classify individuals from certain groups, those individuals will be indexed inadvertently, leaving them vulnerable to identification by the system despite the intended mitigation. This renders the mitigation ineffective for them and creates a disproportionate risk of identification for vulnerable groups.

Crucially, data evaluating demographic biases for instance segmentation and object detection in realistic surveillance scenarios (e.g., low resolution, distant subjects) is currently lacking. While the potential for bias inferred from related tasks warrants serious concern, its precise impact here requires further investigation. Therefore, index exclusion must be applied cautiously, as its effectiveness and fairness depend heavily on the underlying segmentation model's performance and equity.

\subsection{Confusion Loss}
To account for potential weaknesses of index exclusion, we propose a \emph{confusion loss} term as a complementary strategy applied during the AI model's training process, inspired by the MS loss of \citet{wang2019multi}. It encourages the model to produce dissimilar embeddings for images depicting the same person, making it difficult for the system to reliably match and identify them. Specifically, confusion loss pushes embeddings of samples belonging to any person label in the \emph{forbidden set} $\mathcal{F}$ away from each other.

The total loss for this branch is
$$
\mathcal{L} = \mathcal{L}_{\mathrm{MS}} + \lambda_{\mathrm{conf}}\,\mathcal{L}_{\mathrm{conf}},
$$
where $\mathcal{L}_{\mathrm{MS}}$ is defined in \eqref{eq:ms} and
\begin{align}
\mathcal{L}_{\mathrm{conf}} &= \frac{1}{\gamma} \log\Bigl(1+\sum_{(i,j)\in\mathcal{P}_f} e^{\gamma(s_{ij}-m_f)}\Bigr).
\end{align}
Here $s_{ij}=\cos(\mathbf{f}_i,\mathbf{f}_j)$ is the cosine similarity between embeddings $\mathbf{f}_i$ and $\mathbf{f}_j$.
The forbidden pair set is defined as
\begin{equation}
\mathcal{P}_f = \{(i,j):i<j,\,y_i,y_j\in\mathcal{F},\,s_{ij}>m_f\},
\end{equation}
which identifies pairs within the forbidden set that require additional separation.
$\gamma$ sets how sharply the confusion loss focuses on the most similar forbidden pairs, while the margin $m_f$ controls the minimum separation enforced. The weight
$\lambda_{\mathrm{conf}}$ balances the two losses. We compute $\mathcal L_{\mathrm{conf}}$ over all forbidden pairs in the mini-batch.

\section{Experiments}
Our experimental evaluation addresses three key research objectives: (1) demonstrating that generic instance search models develop person re-ID capabilities without exposure to human training data, (2) evaluating the effectiveness of proposed mitigation strategies, and (3) assessing the robustness of these mitigations against circumvention attempts.
To isolate the contribution of learned feature representations from segmentation artifacts, we focus our analysis on Stage 2 of the instance search pipeline (Figure \ref{fig:architecture}). This design choice assumes perfect segmentation masks are available, allowing us to directly evaluate whether embedder models trained exclusively on non-human objects can nonetheless learn to distinguish between individual persons.

We employ a diverse collection of datasets to ensure comprehensive evaluation across different scenarios and difficulty levels:

\begin{itemize}
    \item \textbf{YouTube-VIS} \cite{yang2019video} serves as our training dataset, originally developed for video instance segmentation tasks. We use the 2022 YouTube-VIS training dataset comprising 8,430 unique object instances across 40 categories, including approximately 2,000 unique person identities. We adopt the train-validation-test split and pre-processing pipeline from \citet{10.1145/3617233.3617249}.

    \item \textbf{OVIS} \cite{qi2022occluded} provides an evaluation scenario with heavily occluded objects. The dataset depicts over 5,000 unique objects across 25 object categories, with approximately 800 person identities. Since official test annotations are not released, we use the training set for cross-dataset evaluation. Due to limited computational resources we only evaluate selected models on this larger dataset.

    \item \textbf{CUHK03-NP} \citep{li2014deepreid} is a standard person re-ID benchmark with low-resolution imagery typical of surveillance applications. The dataset includes 700 person identities for evaluation. Following standard practices for this benchmark, we adopt the test split protocol of \citet{zhong2017re}. We use the segmentation masks provided by \citet{song2018mask}.

    \item \textbf{Market-1501} \cite{zheng2015scalable} is another widely used person re-ID benchmark, featuring 751 unique identities and 20,000 gallery images in the official test split. We use the standard evaluation protocol \cite{zheng2015scalable} with segmentation masks from \citet{song2018mask}.
\end{itemize}

\begin{table}
 \vspace{-5pt}
 \centering\resizebox{0.47\textwidth}{!}{%
\begin{tabular}{ l c c c}
 \hline
 & Images & Unique objects & Unique identities \\
 \hline
 Dataset 1 & 92941 / 10330 & 3308 / 375 & 1792 / 206\\
 Dataset 2 &  92954 / 10344 & 3304 / 372 & 0 / 0\\
 \hline
\end{tabular}}
\caption{Training and validation set statistics of the dataset containing all persons in the original YouTube-VIS dataset (Dataset 1), and the dataset where all persons are removed (Dataset 2). Reported numbers correspond to the train/validation sets. Unique identities refers to the number of unique persons in the dataset.
 }
 \label{table:dataset_stats}
\end{table}

\subsection{Demonstrating Overlearned Person Re-ID Capabilities}

\subsubsection{Setup}
We construct two training datasets: (1) one retaining all instances of the person class in the original YouTube-VIS dataset, and (2) one where all these person instances are removed.
To ensure that the amount of training data remains similar between the two datasets, we randomly remove non-person objects from dataset (1) until the number of images is comparable to dataset (2). The dataset statistics can be seen in Table~\ref{table:dataset_stats}.
For evaluation, the query set is composed of all person instances present in the YouTube-VIS test set.
We evaluate the different training strategies using mean Average Precision (mAP), while mAP@r \citep{musgrave2020metric} is used during hyperparameter optimization.

We evaluate four different embedding architectures with comparable parameter counts: CLIP-ViT-B/16 \citep{radford2021learning}, CAFormer \citep{yu2023metaformer}, EVA-02 \citep{fang2023eva}, and ResNet-152 \citep{he2016deep}.
All models produce 512-dimensional embeddings and are trained using the MS loss (Equation \ref{eq:ms}). We employ Bayesian hyperparameter optimization to tune learning rates separately for the feature extractor and embedding layer. Training configurations include:
\begin{itemize}
    \item Batch sizes: 64 (CLIP), 256 (ResNet), 32 (EVA), 16 (CAFormer)
    \item Training duration: 15 epochs (fine-tuning), 30 epochs (from scratch)
    \item Optimizer: AdamW with standard weight decay
\end{itemize}

\subsubsection{Results}
Table 2 and 3 reveal striking evidence of overlearned person re-ID capabilities across all architectures. Models trained without any person data (Dataset 2) achieve remarkably high person re-ID performance:
\begin{itemize}
    \item CLIP maintains 86.8\% mAP without person training (vs. 88.7\% with persons)
    \item All architectures demonstrate substantial person re-ID capabilities despite never observing human subjects during training
\end{itemize}
The minimal performance gap between training regimes suggests that generic object representations naturally encode person-discriminative features.
Cross-dataset evaluation on CUHK03 and Market-1501 (Table \ref{table:w_pretraining}) confirms these capabilities generalize to realistic surveillance scenarios. While performance decreases on these challenging low-resolution datasets, models without person training still achieve 33.8\% mAP on the CUHK-labeled split, demonstrating significant re-ID capabilities.

\begin{table}
 \vspace{-5pt}
  \centering\resizebox{0.47\textwidth}{!}{%
\begin{tabular}{lccccc}
 \hline
 {} & CLIP & CLIP$^*$  & ResNet152 & CAFormer & EVA02\\
 \hline
  {} & \multicolumn{5}{c}{Dataset 1 (with persons)} \\
 \hline
  YVIS & 88.7& 79.7& 82.6& 85.9& 89.7\\
  OVIS & 68.4& -& -& -& -\\
  CUHKlab. & 34.7& 20.1& 26.0& 29.5& 38.3\\
  CUHKdet. & 34.1& 18.6& 25.3& 28.4& 37.2\\
  Market& 26.3& 18.4& 23.7& 25.8& 30.7\\
  \hline
  {} & \multicolumn{5}{c}{Dataset 2 (without persons)} \\
 \hline
  YVIS & 86.8& 78.4& 80.0& 82.5& 87.2\\
  OVIS & 66.9& 57.2& -& -& -\\
  CUHKlab. & 33.8& 19.6& 23.4& 27.5& 34.6\\
  CUHKdet. & 32.8& 18.5& 23.4& 26.1& 33.8\\
  Market & 23.5& 17.7& 21.6& 24.1& 25.8\\
 \hline
\end{tabular}}
 \caption{Person re-ID results (mAP) on all person queries from different datasets for models trained on the version of the YouTube-VIS dataset with persons (Dataset 1) and without persons (Dataset 2).
 CLIP$^*$ corresponds to the evaluation of CLIP trained from scratch.}
 \label{table:w_pretraining}
\end{table}

\begin{table}
 \vspace{-5pt}
  \centering\resizebox{0.47\textwidth}{!}{%
\begin{tabular}{ l c c c}
 \hline
 & Dataset 2& Dataset 2*& Baseline \\
 \hline
  YVIS & 86.8 & 78.4 & 68.3\\
 OVIS & 66.9 & 57.2 & 44.6\\ 
 CUHK03-lab. & 33.8 & 19.6 & 22.9\\ 
 CUHK03-det. & 32.8 & 18.5 & 21.3 \\
 Market-1501 & 23.5 & 17.7 & 14.7\\
 \hline
\end{tabular}}
 \caption{Person re-ID results (mAP) without persons from YouTube-VIS included in training. Results under~$^*$ correspond to models trained from scratch and \textit{Baseline} refers to the pretrained model.}
 \label{table:cuhk}
\end{table}

\subsection{Evaluation of Technical Safeguards}
We now assess technical safeguards intended to curb the identified person re-ID capability.

\subsubsection{Setup: Index Exclusion}
For the index exclusion strategy, we simulate a pipeline where instance segmentation governs indexing, using Grounding Dino \cite{liu2024grounding} as the zero-shot detector on full frames, combined with the Segment Anything Model (SAM) \cite{kirillov2023segment}. We use Grounding Dino due to its state-of-the-art performance on open-vocabulary detection among open-source models, and SAM for its promptable segmentation capabilities, simulating a system designed to identify a wide range of objects. For all evaluation datasets, we prompt the detector to detect the same classes present in YouTube-VIS, except for OVIS where we use the same classes as in the OVIS dataset.

After obtaining detections, we apply class-agnostic Non-Maximum Suppression with an Intersection over Union (IoU) threshold of 0.7 to remove duplicate detections. We then use mask-level IoU combined with Hungarian assignment to establish optimal matching between the detections and ground truth objects. These matches and non-matches are used to categorize the detections into the four cases described below.

For each detection that is ultimately \emph{included} in the index, we use the matched ground-truth mask when computing the embedding.
The inclusion or exclusion of an instance embedding in the final index depends on the detector's output in four ways:

\begin{enumerate} 
    \item \textbf{Detected Person Exclusion}: If Grounding Dino detects an object and classifies it as ``person", and the detected mask significantly overlaps (IoU $>$ 0.5) with the instance's mask, the corresponding embedding is explicitly excluded from the index.
    \item \textbf{Undetected Person Exclusion}: If Grounding Dino fails to detect a ground truth person instance altogether, it is implicitly excluded from the index, simulating scenarios where undetected objects are not processed for embedding.
    \item \textbf{Misclassified Person Inclusion}: If Grounding Dino detects a ground truth person but assigns it an incorrect, non-person label (misclassification), the corresponding embedding is not excluded by this strategy and remains in the index.
    \item \textbf{False Positive Exclusion}: If Grounding Dino detects a non-person object but assigns it an incorrect person label, the corresponding embedding is erroneously excluded, harming non-person retrieval.
\end{enumerate}
The performance of Grounding Dino and SAM, including potential demographic biases as discussed previously, directly influences the frequency and distribution of outcomes 1-4 in our experiments.
Case 3 undermines privacy protection, whereas Case 4 diminishes utility.

\subsubsection{Setup: Confusion Loss}
We apply multi-objective Bayesian optimization to find the values of $\lambda_{\mathrm{conf}}$, $m_f$, and the learning rate that simultaneously (i) minimize person mAP@R and (ii) maximize mAP@R on non-person classes. Pre-trained CLIP-ViT-B/16 is fine-tuned for a single epoch (longer schedules increased person mAP) on the full YouTube-VIS split that still contains persons. We use the AdamW optimizer with standard weight decay.

\subsubsection{Results}
\begin{figure*}
    \centering
    \begin{subfigure}[b]{0.95\textwidth}
        \includegraphics[width=\textwidth]{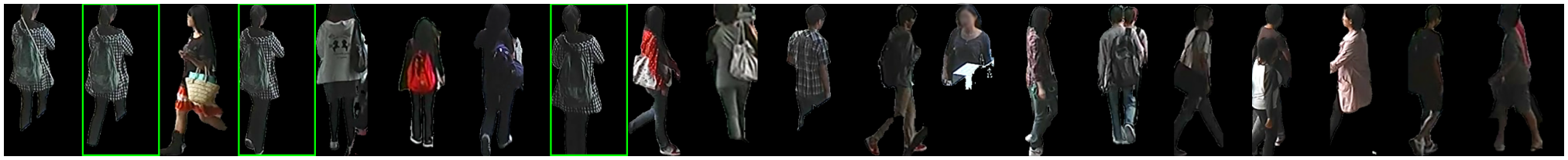}    
        \includegraphics[width=\textwidth]{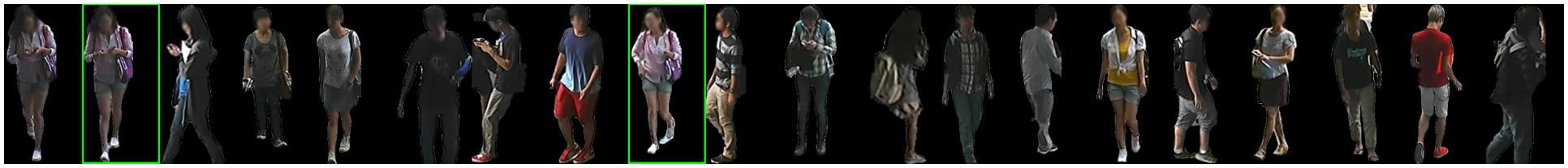}
        \includegraphics[width=\textwidth]{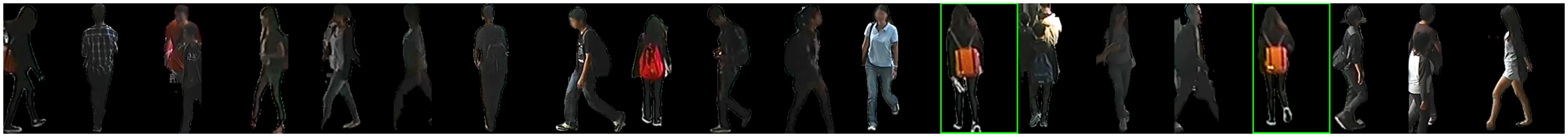}
         \includegraphics[width=\textwidth]{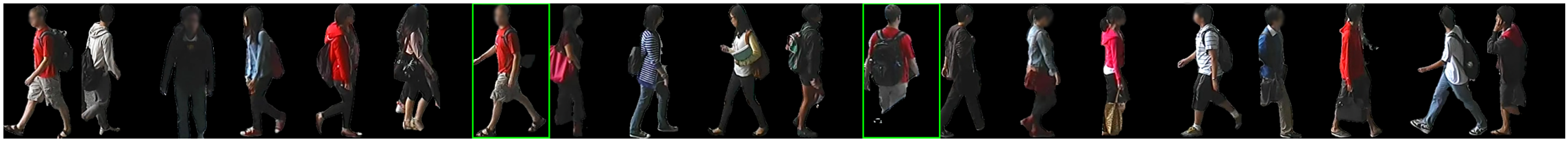}
        \includegraphics[width=\textwidth]{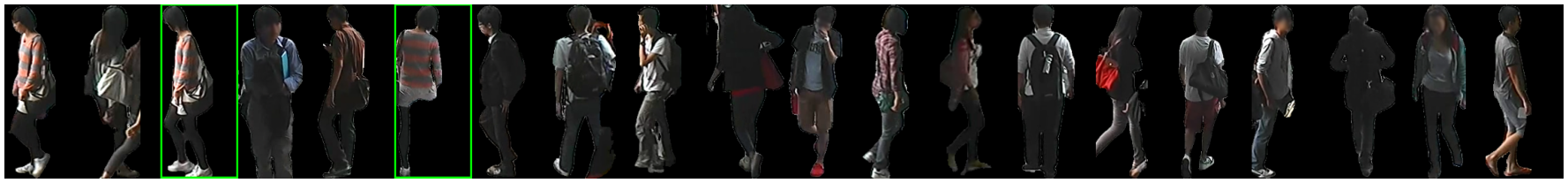}
    \end{subfigure}
    \caption{Examples where the confusion-loss model fails to prevent re-identification. The query appears in the left-most column; green borders denote correct matches. The visible faces are blurred for this figure.}
    \label{fig:confusion_fails}
\end{figure*}

\begin{figure}
    \centering
    \begin{subfigure}[b]{0.29\textwidth}
        \centering
        \includegraphics[width=\textwidth]{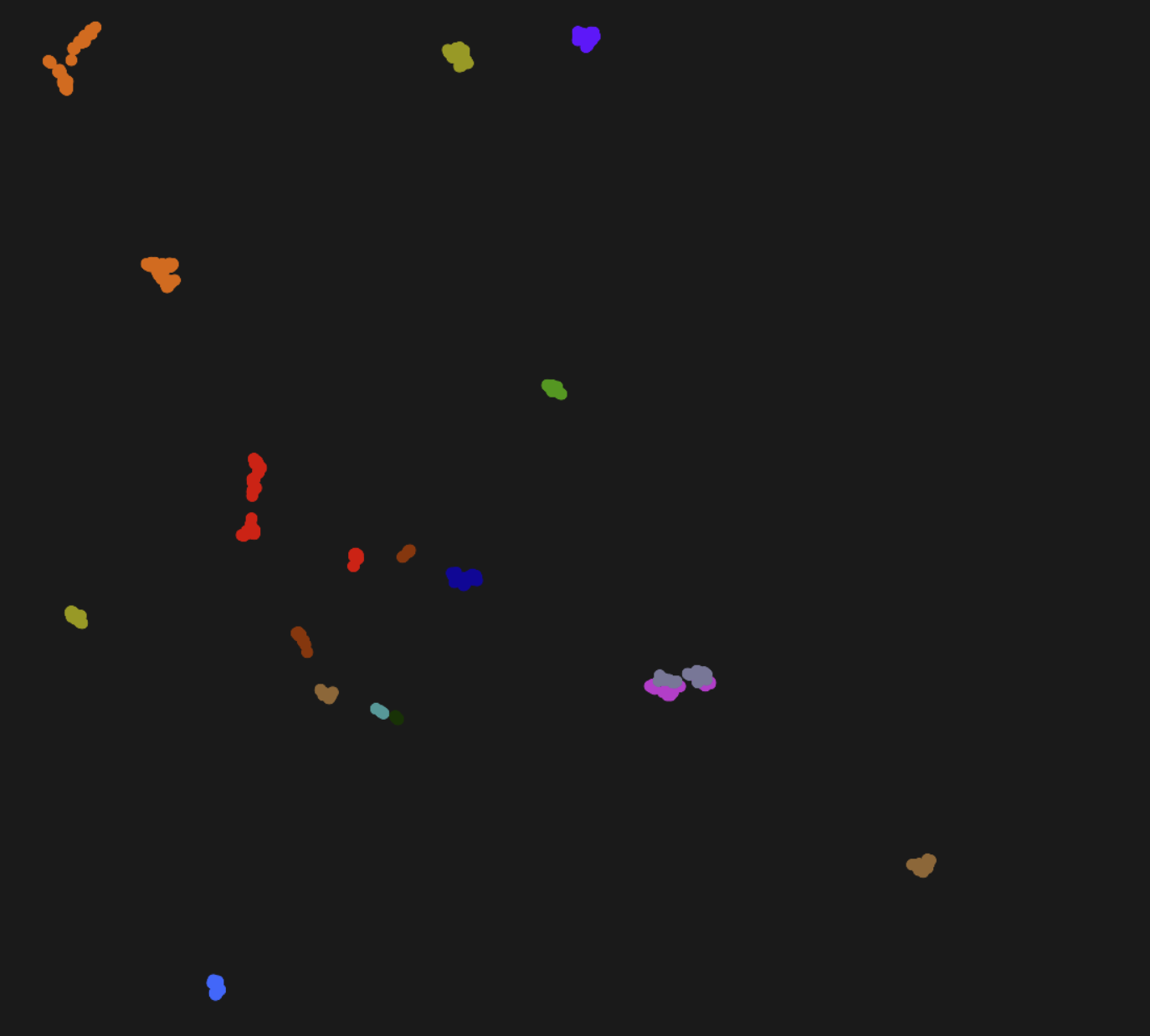}
        \caption{MS loss}
        \label{fig:center}
    \end{subfigure}
    \hfill
    \begin{subfigure}[b]{0.29\textwidth}
        \centering
        \includegraphics[width=\textwidth]{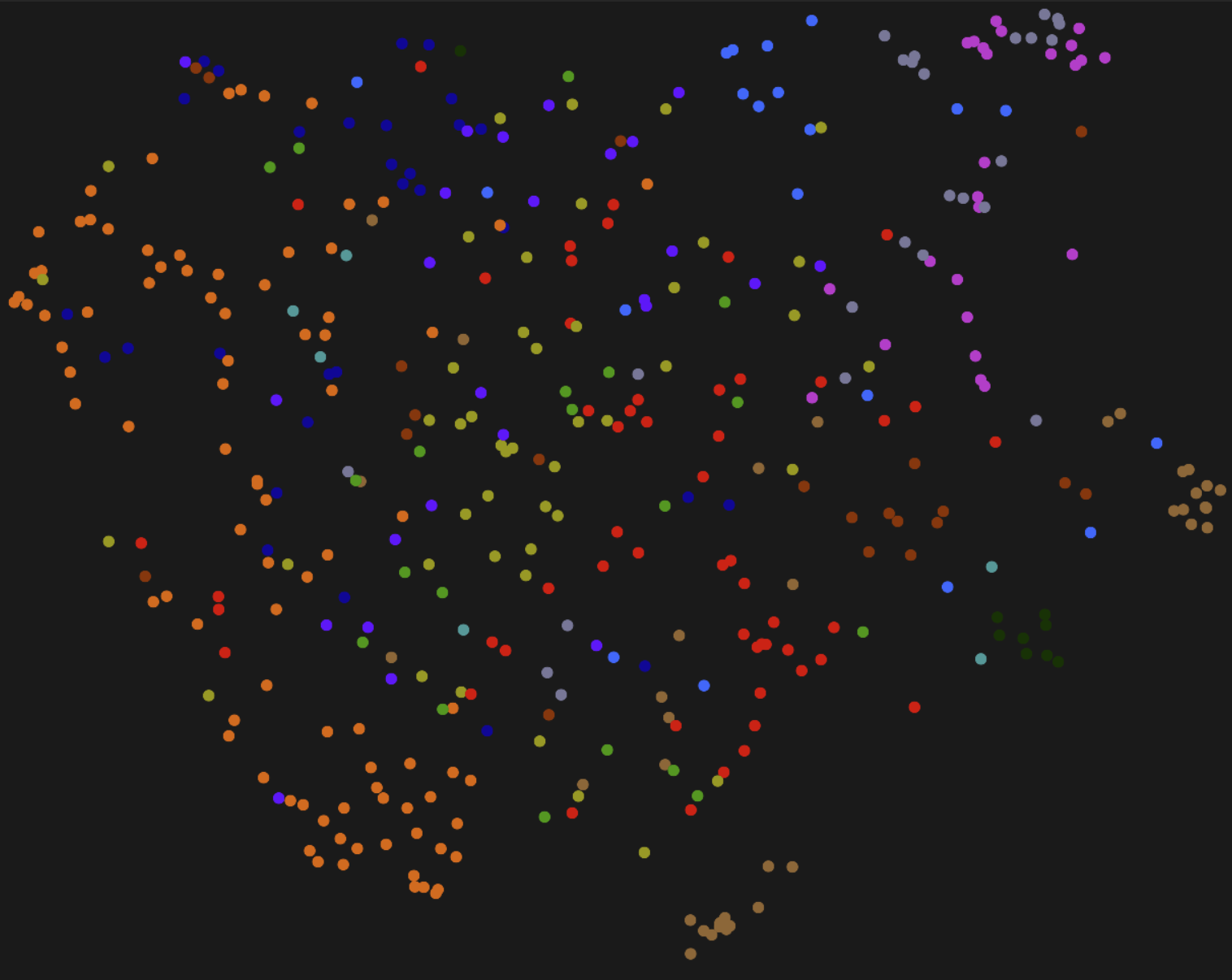}
        \caption{Confusion loss}
        \label{fig:confusion}
    \end{subfigure}
    \caption{UMAP visualization comparing YouTube-VIS embeddings from models trained with (a) standard MS loss on non-human subjects and (b) confusion loss. Colors represent different individuals; confusion loss effectively disperses person embeddings, reducing re-identification.
    }
    \label{fig:umap}
\end{figure}

\begin{figure}[htbp]
    \centering
    \begin{minipage}[b]{0.27\columnwidth}
        \centering
        \textbf{Original}
    \end{minipage}%
    \begin{minipage}[b]{0.27\columnwidth}
        \centering
        \textbf{MS}
    \end{minipage}%
    \begin{minipage}[b]{0.27\columnwidth}
        \centering
        \textbf{Confusion}
    \end{minipage}\\[2mm]

    \begin{subfigure}[b]{0.27\columnwidth}
        \includegraphics[width=\textwidth]{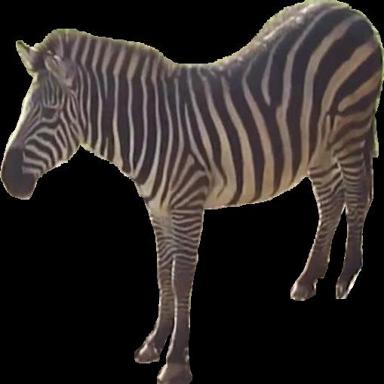}
    \end{subfigure}%
    \begin{subfigure}[b]{0.27\columnwidth}
        \includegraphics[width=\textwidth]{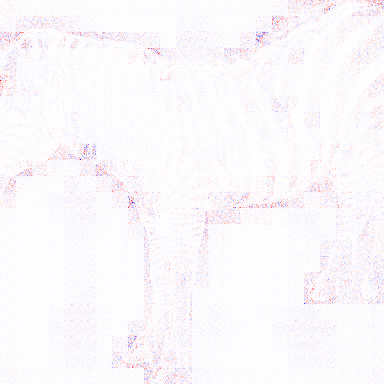}
    \end{subfigure}%
    \begin{subfigure}[b]{0.27\columnwidth}
        \includegraphics[width=\textwidth]{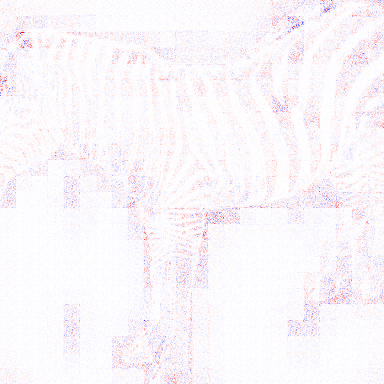}
    \end{subfigure}\\[1mm]
    \begin{subfigure}[b]{0.27\columnwidth}
        \includegraphics[width=\textwidth]{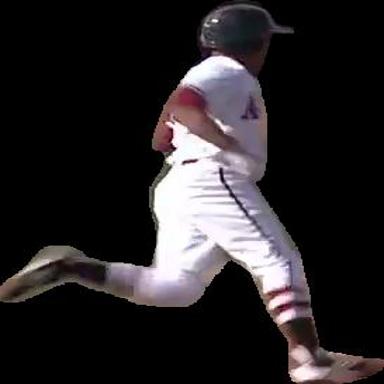}
    \end{subfigure}%
    \begin{subfigure}[b]{0.27\columnwidth}
        \includegraphics[width=\textwidth]{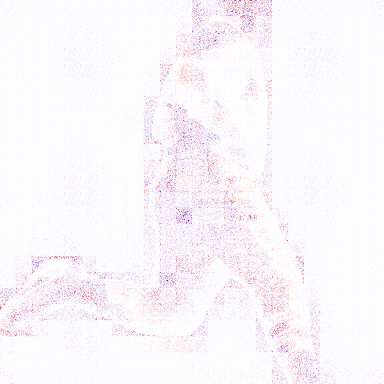}
    \end{subfigure}%
    \begin{subfigure}[b]{0.27\columnwidth}
        \includegraphics[width=\textwidth]{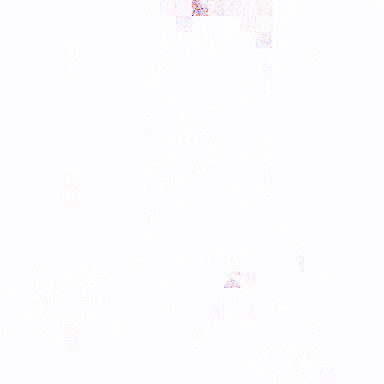}
    \end{subfigure}\\[1mm]
    \begin{subfigure}[b]{0.27\columnwidth}
        \includegraphics[width=\textwidth]{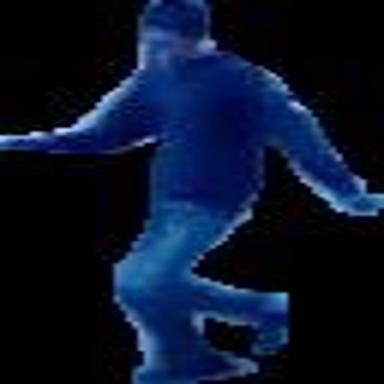}
    \end{subfigure}%
    \begin{subfigure}[b]{0.27\columnwidth}
        \includegraphics[width=\textwidth]{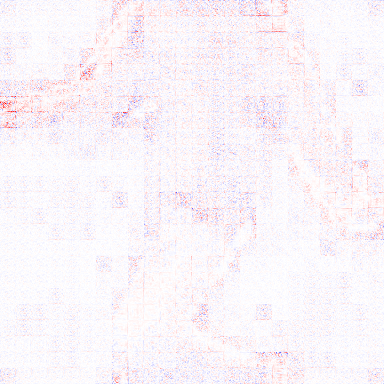}
    \end{subfigure}%
    \begin{subfigure}[b]{0.27\columnwidth}
        \includegraphics[width=\textwidth]{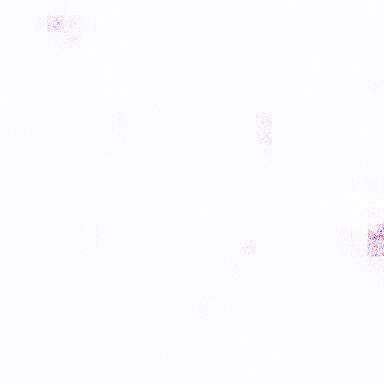}
    \end{subfigure}\\[1mm]
    \begin{subfigure}[b]{0.27\columnwidth}
        \includegraphics[width=\textwidth]{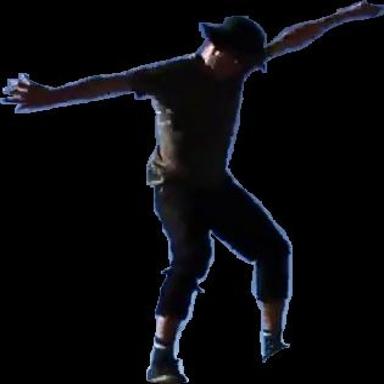}
    \end{subfigure}%
    \begin{subfigure}[b]{0.27\columnwidth}
        \includegraphics[width=\textwidth]{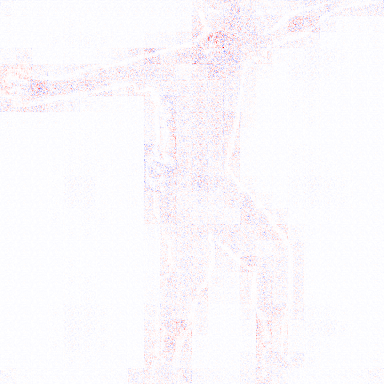}
    \end{subfigure}%
    \begin{subfigure}[b]{0.27\columnwidth}
        \includegraphics[width=\textwidth]{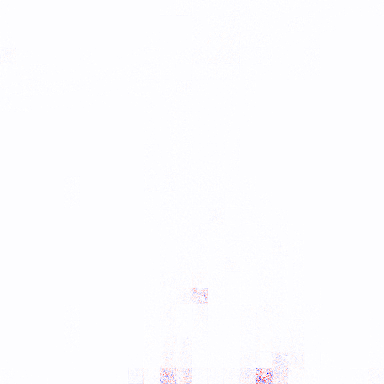}
    \end{subfigure}

    \caption{AttnLRP relevance heatmaps comparing models trained with standard MS loss versus confusion loss. Confusion loss reduces the relevance of identifiable person features, while preserving relevance for general object recognition.
    }
    \label{fig:attention}
\end{figure}

Table \ref{table:mitigation} demonstrates the effectiveness of our mitigation strategies:
\begin{itemize}
    \item \textbf{Index Exclusion} nearly eliminates person re-identification (4.4\% mAP on YouTube-VIS) while preserving 89.0\% of non-person retrieval performance. However, real-world performance may depend on potential demographic biases.
    \item \textbf{Confusion Loss} reduces person re-ID to 10.6\% mAP but incurs a higher utility cost (7.1 percentage point drop in non-person retrieval). The confusion loss effectively disperses person embeddings in feature space, as visualized in Figure \ref{fig:umap}. 
    \item \textbf{Combined approach} achieves the strongest privacy protection (1.7\% person mAP) by capturing persons that evade initial detection. The combined utility cost remains acceptable (7.9 percentage point reduction). 
\end{itemize}
Figure \ref{fig:confusion_fails} shows retrieval results of cases where the confusion loss-trained model fails to prevent re-identification. 
Figure \ref{fig:attention} shows AttnLRP heatmaps \cite{achtibat2024attnlrp} from the model trained with regular MS loss, and the model trained with confusion loss. Relevance was propagated from a single scalar obtained by taking the L2 norm of the embedding, allowing AttnLRP to attribute pixels that most strongly increase the magnitude of the representation. We consistently find that while the confusion loss-trained model assigns minimal relevance to identifying features of person images (rows 2-4), it correctly attributes high relevance to discriminative features of non-person images (first row).

\begin{table}
  \vspace{-5pt}
  \centering\resizebox{0.47\textwidth}{!}{%
\begin{tabular}{lcccc}
 \hline
 & MS & Ind. & Conf. & Ind.+Conf.\\
 \hline
 YVIS & 86.8 / 89.9 & 4.4 / 89.0 & 10.6 / 82.8 & 1.7 / 82.0 \\
 OVIS & 66.9 / 63.5 & 0.8 / 63.0 & 7.7 / 51.9 & 0.1 / 51.5 \\
 CUHK03-lab. & 33.8 & 0.9 & 0.6 & 0.2 \\
 CUHK03-det. & 32.8 & 0.8 & 0.5 & 0.1 \\
 Market-1501 & 23.5 & 3.7 & 0.4 & 0.2 \\
 \hline
\end{tabular}}
 \caption{Test-set mAP for standard multi-similarity (MS) loss and each mitigation strategy: Ind. for index exclusion and Conf. for confusion loss. Results are reported for person queries and for non-person queries on YouTube-VIS and OVIS (i.e.,  person / no person).}
 \label{table:mitigation}
\end{table}

\subsection{Robustness and Vulnerability Analysis}
We next demonstrate that the confusion loss filter can be partially circumvented by querying with smaller, less person-like regions.

\subsubsection{Setup}
Grounding DINO combined with SAM extracts person parts, i.e. upper clothing and bags, for use as queries and references. Additionally, we crop a central window from each person image, removing 40\% of the height from the top and 57.5\% from the bottom.

\subsubsection{Results}
Table \ref{table:robustness} reveals vulnerabilities in confusion loss protection: cropping restores person re-ID to 6.1\% mAP, bag-based queries achieve 4.1\% mAP, and upper clothing enables 3.0\% mAP.
Improved segmentation models, especially for accessories, may amplify this vulnerability. Moreover, the results reveal that instance search models trained with standard MS loss can achieve significant person re-ID through accessories and clothing alone.

\begin{table}
 \vspace{-5pt}
\centering\resizebox{0.35\textwidth}{!}{%
\begin{tabular}{ l c c }
 \hline
 & MS & Confusion loss \\
 \hline
 Cropped & 13.0 & 6.1 \\
 Bags & 18.5& 4.1\\
 Upper clothing & 25.6 & 3.0 \\
\end{tabular}}
 \caption{CUHK03-labeled mAP when queries and references are restricted to smaller regions of the original person images.}
 \label{table:robustness}
\end{table}

These findings imply that queries which do not obviously depict humans can evade the confusion loss filter. They also raise the open question of whether certain religious or cultural garments, under-represented in the training set, could slip through more easily. Overall, state-of-the-art detectors suffice for index exclusion guardrails, and pairing index exclusion with confusion loss appears promising --- yet comprehensive demographic testing on realistic CCTV footage remains essential.

\section{Legal and Policy Implications}
Our findings, demonstrating overlearned person re-ID capabilities in generic instance search models, present a direct challenge to the data protection regime in Europe. Further research must also examine such systems in the context of the new AI regulation in Europe. The AI Act classifies as high-risk and extensively regulates the use of AI systems for biometric identification and categorization or profiling \cite[Art.6 in conj. with Annex III (6) and Chapter III]{AIAct_2021}. However, from the analysis above we demonstrated that the use of biometrics or sensitive data is not necessary in models with sophisticated feature extraction in order to single out people. Consequently, instance search algorithms provide many identification capabilities that may fall out of the scope of the legal definitions of biometrics, sensitive data, or profiling \ref{sec:background} and therefore also from the high-risk AI systems regime. The capacity of generic instance search models to re-identify individuals, even without explicit training on human data, significantly strains established data protection principles, particularly purpose limitation and data minimization. If a system deployed for general object search can, through overlearning, effectively single out individuals, its operation may constitute processing of personal data (potentially sensitive data) beyond its original, legitimate purpose. 
This exemplifies ``function creep", where a technology's capabilities and uses imperceptibly expand beyond its intended scope \cite{koops_concept_2021}.
Further, the AI Act embraces a risk-based logic yet leaves a gray zone for models whose architecture fosters person-level matching without any explicit biometric design. 

 The emergence of re-ID in AI systems raises critical questions like:

\begin{itemize}
    \item At what threshold of accuracy or reliability does an overlearned re-ID capability transform a generic AI tool into a de facto high-risk biometric identification system under the AI Act?
    \item How can AI policies effectively address systems that acquire sensitive capabilities, like person re-ID, not through explicit design but through the inherent learning processes of complex models?
    \item How can organizations ensure compliance with purpose limitation when AI systems develop unforeseen, privacy-invasive capabilities post-deployment? It also remains unclear what constitutes adequate de-identification for learning algorithms.
\end{itemize}

While an extensive legal analysis is left for future work, in our view, the regulatory focus on AI biometric identification leaves in a legal gray area other AI identification capabilities that are discovered and potentially utilized, even if not initially designed. Pre-market conformity and post-market monitoring of AI systems should include clear benchmarks for the early detection of overlearning potential post-deployment via auditing or ``red teaming". While technical mitigations offer partial solutions, their vulnerabilities underscore the need for clear testing and certification standards on emergent identification capabilities to avoid function creep, privacy violations, and excessive surveillance.

\section{Conclusion}

Our research demonstrates that generic instance search models develop significant person re-identification capabilities even when trained exclusively on datasets without individuals. This overlearned capability raises important data protection and regulatory concerns, particularly as these systems may fall outside the scope of current biometric identification regulations despite functionally enabling similar surveillance capabilities. While our evaluation of technical mitigations shows that the combination of index exclusion and confusion loss can effectively reduce person re-ID accuracy while maintaining reasonable performance on non-person retrieval tasks, these solutions remain vulnerable to circumvention through partial person queries. 
Future work should pursue more robust mitigation strategies resilient to circumvention attempts, explore the effectiveness of these mitigations across diverse demographic groups in realistic surveillance scenarios, and develop methodology for proactively identifying overlearned capabilities before deployment. As AI systems continue to grow in complexity and capability, our work highlights the necessity of expanding technical and regulatory approaches to address not just the intended functions of AI systems but also their potential overlearned capabilities, particularly when these capabilities may impact fundamental rights such as privacy and non-discrimination.

\section{Acknowledgments}
This work was supported by industry partners and the Research Council of Norway (project number 320783). 
We thank SFI NORCICS for providing computational resources.

\bibliography{aaai25}

\end{document}